\documentclass[letterpaper, 10 pt, conference]{ieeeconf}  %

\IEEEoverridecommandlockouts                              %

\usepackage[table]{xcolor}
\usepackage{graphicx}
\usepackage{cite}
\usepackage{multirow}
\usepackage{booktabs}
\usepackage{amsmath}
\usepackage{amssymb}
\usepackage{microtype}

\usepackage[utf8]{inputenc}
\usepackage[T1]{fontenc}
\usepackage{hyperref}

\title{\LARGE \bf
Enhancing Visual Place Recognition via\\Fast and Slow Adaptive Biasing in Event Cameras
}

\author{Gokul B. Nair \qquad\qquad Michael Milford \qquad\qquad Tobias Fischer %
\thanks{This work received funding from Intel Labs to TF and MM, and from ARC Laureate Fellowship FL210100156 to MM. The authors acknowledge continued support from the Queensland University of Technology (QUT) through the Centre for Robotics and the Research Engineering Facility (REF) team for providing engineering support, expertise and research infrastructure.}%
\thanks{The authors are with the Centre for Robotics, Faculty of Engineering, Queensland University of Technology, Brisbane, QLD Australia 4000
{\tt\small gokulbnr@gmail.com}}%
}

\usepackage{eso-pic}
\usepackage{url}
\AddToShipoutPicture*{%
     \AtTextUpperLeft{%
         \put(-3.5,10){
           \begin{minipage}{\textwidth}
              \scriptsize
              \MakeUppercase{Preprint version | Paper accepted to the IEEE/RSJ International Conference on Intelligent Robots and Systems (IROS 2024)} 
           \end{minipage}}%
     }%
}

\begin{document}

\bstctlcite{bstctl:forced_etal,bstctl:nodash}

\maketitle
\thispagestyle{empty}
\pagestyle{empty}

\begin{abstract}
Event cameras are increasingly popular in robotics due to beneficial features such as low latency, energy efficiency, and high dynamic range. Nevertheless, their downstream task performance is greatly influenced by the optimization of bias parameters. These parameters, for instance, regulate the necessary change in light intensity to trigger an event, which in turn depends on factors such as the environment lighting and camera motion. This paper introduces feedback control algorithms that automatically tune the bias parameters through two interacting methods: 1) An immediate, on-the-fly \textit{fast} adaptation of the refractory period, which sets the minimum interval between consecutive events, and 2) if the event rate exceeds the specified bounds even after changing the refractory period repeatedly, the controller adapts the pixel bandwidth and event thresholds, which stabilizes after a short period of noise events across all pixels (\textit{slow} adaptation). Our evaluation focuses on the visual place recognition task, where incoming query images are compared to a given reference database. We conducted comprehensive evaluations of our algorithms' adaptive feedback control in real-time. To do so, we collected the QCR-Fast-and-Slow dataset that contains DAVIS346 event camera streams from 366 repeated traversals of a Scout Mini robot navigating through a 100 meter long indoor lab setting (totaling over 35km distance traveled) in varying brightness conditions with ground truth location information. Our proposed feedback controllers result in superior performance when compared to the standard bias settings and prior feedback control methods. Our findings also detail the impact of bias adjustments on task performance and feature ablation studies on the fast and slow adaptation mechanisms.

\end{abstract}
\section{Introduction}
\label{sec:introduction}

Event cameras distinguish themselves from traditional cameras by their unique operational principle. Unlike traditional cameras, which capture images at set intervals, event cameras generate data when individual pixels detect changes in brightness, leading to OFF or ON events when the change falls below or exceeds a certain threshold. The resulting output is an event stream that includes the $(x,\, y)$ pixel coordinates, a high-precision timestamp $t$, and the polarity $p$ indicating the event type (OFF/ON). This sparse and asynchronous data capture method grants event cameras high temporal resolution, low latencies, and reduced power consumption; qualities that could be advantageous for autonomous systems, particularly in tasks like motion segmentation, pose estimation, visual odometry, simultaneous localization and mapping (SLAM) and neuromorphic control~\cite{gallego2020event}.

\begin{figure}[t]
  \centering
  \vspace{1.7mm}
  \includegraphics[width=1\linewidth]{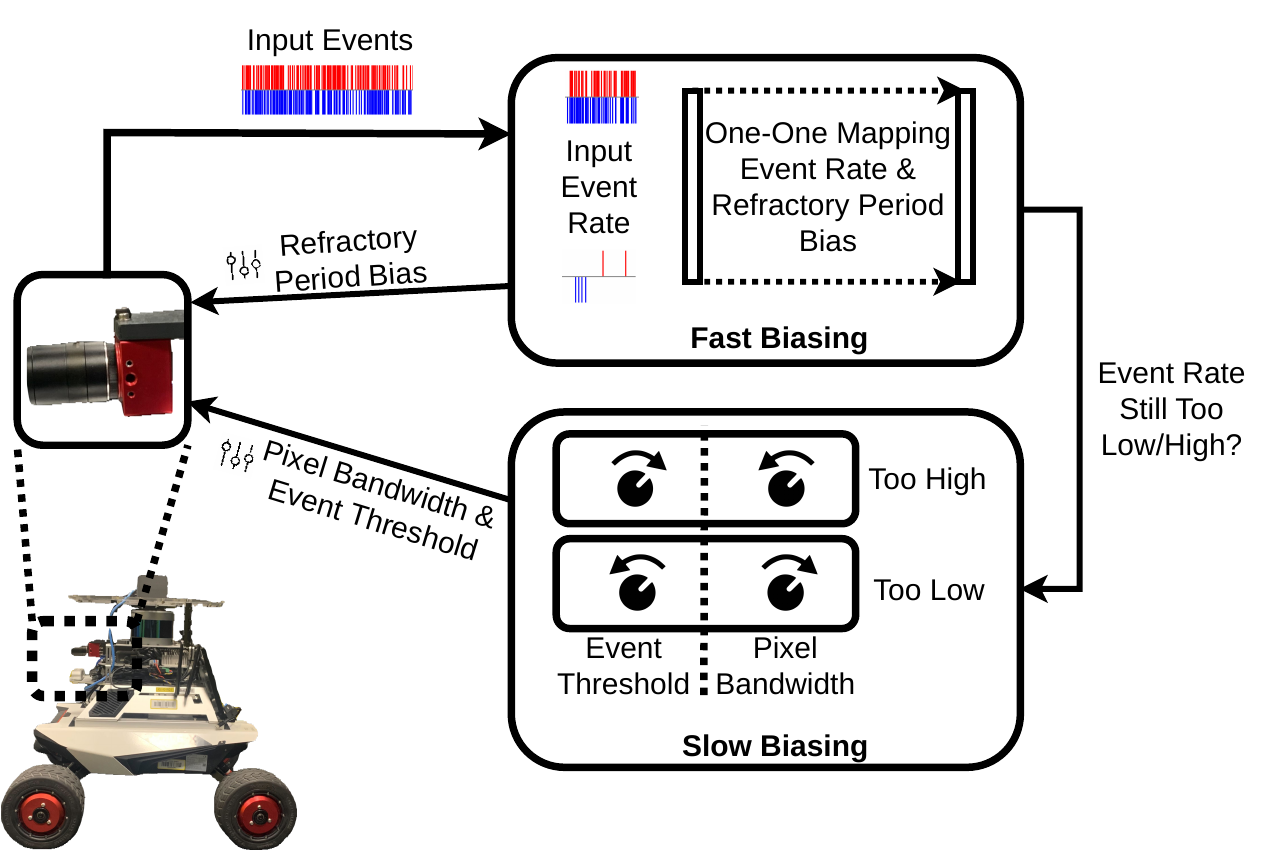}
      \caption{\textbf{Closed-loop biasing of event cameras.} We perform feedback control on event camera biases to obtain improved performance for the visual place recognition downstream task. We achieve this by varying refractory period, pixel bandwidth, and event threshold parameters in real-time. As illustrated, we modulate these device parameters to monitor and maintain event-rate from an onboard event-camera mounted on a mobile robot.}
      \vspace*{-0.2cm}
  \label{fig:teaser}
\end{figure}

While traditional camera performance has greatly benefited from extensive optimization efforts over the years, such as adjustments to converter gain, noise filtering, and exposure time, this paper focuses on the development of similar automated feedback control strategies for event cameras. Given that it is often impractical to manually adjust an event camera's bias parameters during robotic missions, this study presents a case for the necessity of intelligent, automated tuning based on the event rate -- a metric that counts the number of events within a specified \mbox{timeframe -- to} significantly enhance performance in downstream tasks (Figure~\ref{fig:teaser}).

We demonstrate our proposed feedback controller on the Visual Place Recognition (VPR) task~\cite{garg2021your,schubert2023visual}. VPR benefits the long-term autonomy of mobile robots as a core component in localization and mapping algorithms, aiding loop-closure in SLAM and global re-localization in kidnapped robot scenarios. Event-based VPR has seen an increased interest over the past few years~\cite{fischer2022many,kong2022event,fischer2020event,lee2021eventvlad,lee2023ev,yu2023brain,zhu2023neuromorphic,hou2023fe}, but an in-depth investigation of the impact of the bias parameters on VPR performance is lacking until now.
Our contributions are summarized as follows: 
\begin{enumerate}
    \item The introduction of a novel closed-loop feedback controller that, for the first time, seamlessly integrates continuous `fast' adjustments of the refractory period with `slow' periodic modifications of bias parameters that influence the camera's sensitivity.
    \item An in-depth analysis of bias parameter tuning for VPR, showcasing the significant performance improvements offered by our fast-and-slow feedback controller over both the standard bias settings and previous feedback control methods \cite{delbruck2021feedback}.
    \item We collect and make publicly available the QCR-Fast-and-Slow-Dataset which is used to compare and evaluate our fast-slow feedback controller with prior approaches. The dataset contains 366 traverses of an indoor lab environment (100 meters) across high, medium, and low brightness conditions spanning various combinations of bias parameters. The dataset contains the event streams, conventional image data, and ground truth robot poses.
\end{enumerate}

To support ongoing research in this area, we release both the algorithm and the QCR-Fast-and-Slow Dataset: \href{https://github.com/gokulbnr/fast-slow-biased-event-vpr}{https://github.com/gokulbnr/fast-slow-biased-event-vpr}

\section{Related Works}
\label{sec:relatedworks}
This section first provides an overview of typical approaches to process an event stream in Section~\ref{subsec:dvsprocessing}. We then introduce event-based VPR frameworks in Section~\ref{subsec:vpr}. We also note that~\cite{gallego2020event} provides an excellent overview of event-based vision, with a focus on applications and algorithms that most benefit from the unique properties of event cameras.

\subsection{Event stream representations and processing}
\label{subsec:dvsprocessing}
As single events do not carry sufficient information for downstream tasks, they are often processed in batches \cite{gallego2020event}. Many early works relied on temporal collections of events that are processed into image-like event frames, where each pixel contains information about e.g.~the count of events within a time window~\cite{rebecq2016evo, maqueda2018event}. Such batching enables the processing of these event frames with standard frame-based computer vision algorithms. The time window can be variable, with some approaches introducing adaptive time-slices of events based on slice event counts to perform feedback control~\cite{liu2018adaptive}. Similarly, \cite{nunes2023adaptive} couples event rates to global scene dynamics to adaptively determine the decay of events. 

When processing event streams in real-time, the high temporal resolution requires algorithms that can process events rapidly. It is desirable to keep the event rate within certain bounds, as higher event-rates can lead to processing delays. \cite{finateu20205} maintains desirable event-rates through various event-dropping strategies at the hardware level. \cite{tapia2022asap} tackles process overflow due to high event rates by modulating event package sizes and controlled discarding of events depending on expectations of process times and hardware limitations. \cite{glover2018controlled} uses a gain parameter to control the absolute delay of the system. 

Tackling process overload through controlled discarding of events leads to task-relevant data loss. \cite{s_2022} adapts ON/OFF event thresholds in the device hardware to monitor and modulate event rates. The popular jAER software~\cite{delbruck2008frame} exposes bias parameters along with underlying coarse and fine bias settings \cite{yang2012addressable}. This opens up pathways to software-triggered adaptive biasing approaches, which couples various information sources to infer environmental conditions and tune the device on the fly.

The most relevant prior work is \cite{delbruck2021feedback}. They perform feedback control to regulate event-rates and noise by changing event threshold, pixel bandwidth, and refractory period. Section~\ref{subsec:baselines} contains a detailed explanation of the various feedback controllers introduced in~\cite{delbruck2021feedback}. 
\cite{dilmaghani2023control} evaluates how changes in device biases affect the output of a Prophesee event camera, using sharpness to assess event stream quality. \cite{mcreynolds2022experimental} characterizes DVS performances to a set of quantifiable parameters and evaluates DVS responses based on parameters such as threshold, bandwidth, refractory period and scene illumination.

\subsection{Event-based Visual Place Recognition}
\label{subsec:vpr}

Visual Place Recognition (VPR) involves recognizing if current sensory data correspond to a previously visited location. This plays a crucial role in localization and navigation capabilities of mobile robotic systems \cite{schubert2023visual}. Typically, VPR pipelines infer geo-locations of incoming query images by matching them with a set of geo-stamped reference images of known places. This addresses the revisiting problem and aids in loop-closures, making VPR a useful component of SLAM systems \cite{engelson1992error}.

Consequently, VPR is an interesting downstream task for event-based sensing, and well suited for our experiments validating the effectiveness of our fast-and-slow feedback controller. In the remainder of this section, we provide an overview of prior works on event-based VPR.

\cite{fischer2020event} combines temporal windows of varying lengths through parallelized ensembles to obtain a high-performing VPR system at the cost of high computational requirements. \cite{lee2021eventvlad} uses the popular Vector of Locally Aggregated Descriptors (VLAD) pooling mechanism to obtain image descriptors after reconstructing event edges. This idea has been taken further in~\cite{kong2022event} which introduced the first end-to-end deep-learning event-based VPR framework. \cite{fischer2022many} introduced a computationally efficient technique that relies on just a small subset of informative pixels. Another interesting approach is the combination of conventional frames with event streams for improved performance~\cite{hou2023fe}.

Moving towards more bio-inspired approaches, \cite{yu2023brain} developed a brain-inspired multimodal hybrid neural network that seeks to imitate the neurological processes of the brain for more efficient robot place recognition. Their model demonstrates the synergy between different sensory modalities and computational paradigms, offering insights into the development of advanced robotic systems. This approach is similar to \cite{lee2023ev} who also used a spiking neural network for processing the event stream. Similarly, \cite{zhu2023neuromorphic} harnessed the power of neuromorphic computing, presenting a neuromorphic sequence learning model that uses event cameras to navigate through vegetative environments.

There are relatively few datasets publicly available that are suited for event-based VPR. The Brisbane-Event-Dataset \cite{fischer2020event} contains 6 traverses with varying time-of-day of an 8km long route captured with a DAVIS346 camera mounted on the windshield of a car. It contains pseudo ground-truth from a GPS device. The QCR-Event-Dataset \cite{fischer2022many} contains 16 traverses of a 160m long route captured with a DAVIS346 camera mounted on a Clearpath Jackal robot, whereby the mounting position was varied between forward-, side-, or downward-facing. The ground truth for a subset of these traverses was manually annotated. The DDD20 dataset \cite{hu2020ddd20} is an enormous dataset containing recordings of over 4000km of driving (51h). It was mainly recorded for the steering prediction task, but a subset of the recordings could be used for the visual place recognition task. To the best of our knowledge, our QCR-Fast-and-Slow dataset is the largest indoor event-based VPR dataset to date (recordings totaling over 35km).

\section{Methodology}
\label{sec:approach}
\begin{figure}[t]
  \centering
  \includegraphics[width=1.0\linewidth]{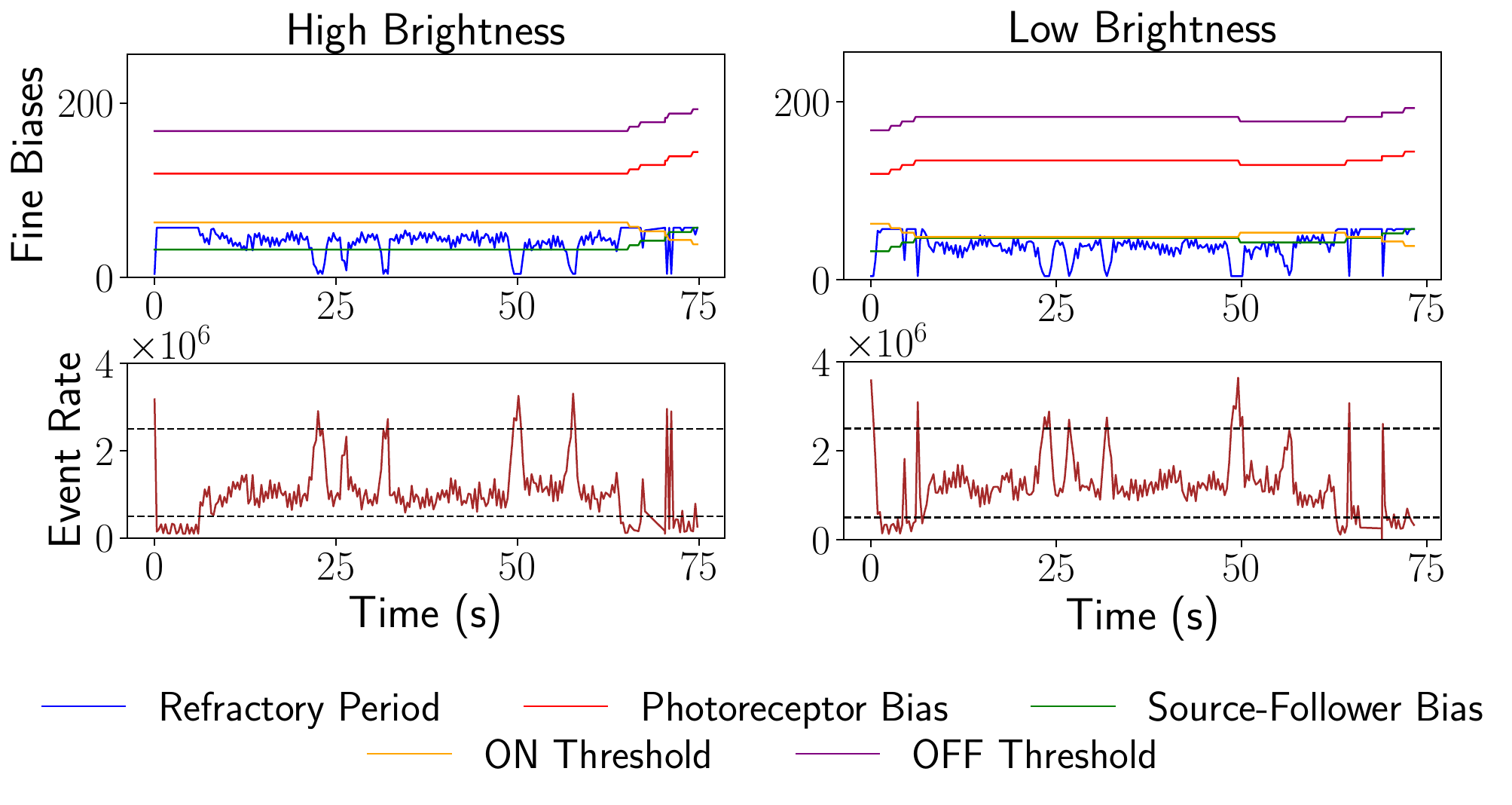}
  \vspace*{-0.5cm}
  \caption{\textbf{Dynamic bias adjustments for differing brightness conditions.} The top two graphs display the fine-tuning of bias parameters over time, while the bottom graphs show the corresponding event rates. The refractory period (blue) is constantly adjusted, while the other bias parameters are only adjusted if the refractory period changes alone do not suffice. On the left, under high brightness conditions, the biases are adjusted less frequently, resulting in a relatively stable event rate. Conversely, on the right, for low brightness conditions, there is more frequent fine-tuning of biases, which correlates with the more variable event rate observed. This illustrates how bias parameter management is critical for adapting to different luminosity levels and maintaining consistent event detection rates.}
  \vspace*{-0.2cm}
  \label{fig:APSvariations1}
\end{figure}

We outline the foundational principles of event cameras for completeness in Section~\ref{subsec:preliminaries}. Our fast-and-slow feedback controller consists of two components that interact with each other (Figures~\ref{fig:teaser} and~\ref{fig:APSvariations1}): a fast controller that changes the refractory period aiming to keep the event rate within specified bounds (Section~\ref{subsec:rate_for_refr}), and a slow controller that changes the pixel bandwidth and event threshold if these event rate bounds are repeatedly exceeded (Section~\ref{subsec:refr_for_biases}).
\subsection{Preliminaries: Event Camera Principles}
\label{subsec:preliminaries}
 Event cameras, or Dynamic Vision Sensors (DVS), capture the visual world differently from conventional cameras. They produce asynchronous events at the pixel level, triggered by changes in brightness that surpass a predefined threshold~\cite{event-based-neuromorphic-systems}. Each event $\mathbf{e}=(x, y, t, p) \in \mathcal{E}$ comprises the pixel coordinates $(x, y)\in \mathbb{N}^{2}$, a high-resolution timestamp $t \in \mathbb{R}^+$, and a polarity $p \in \{-1, 1\}$ indicating the direction of brightness change~\cite{gallego2020event}.

Ideally, event cameras would only generate events upon changes in the scene. However, in practice, they also produce noise events due to factors like junction leakage and parasitic photocurrents in bright conditions, and shot noise in low light \cite{graca2021unraveling, graca2023optimal}. These unwanted events contribute to background activity, diminishing the signal-to-noise ratio \cite{taverni2018front, nozaki2017temperature, guo2022low}.

Post-processing techniques exist to mitigate background noise \cite{mcreynolds2023exploiting, guo2022low}, but on-chip bias current generators also play a crucial role in adapting the camera to varying lighting conditions \cite{delbruck2005bias, delbruck201032, yang2012addressable}. As detailed in the next section, our methodology automates the adjustment of these bias parameters, akin to tuning converter gain, exposure time, and ISO in traditional cameras.

\begin{figure}[t]
    \centering
    \includegraphics[width=0.49\linewidth]{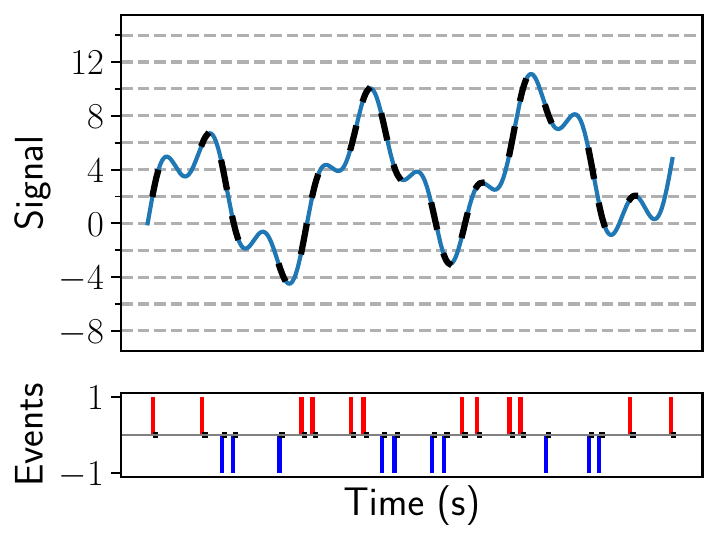}
    \includegraphics[width=0.49\linewidth]{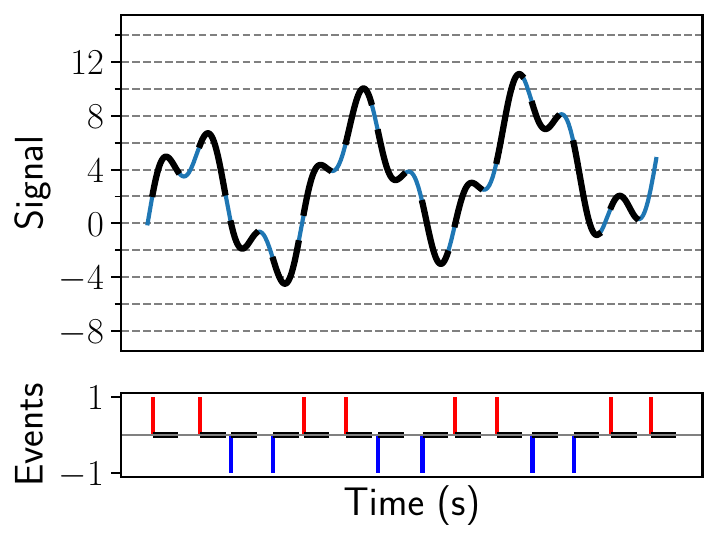}
    \vspace*{-0.3cm}
    \caption{\textbf{Refractory period.} Left: A short refractory period (black lines indicate the refractory period) allows more frequent firing of events (as indicated by the red and blue lines in the bottom plots), resulting in a signal that closely follows the input stimulus. Right: An extended refractory period limits the firing rate, leading to fewer events and a signal that may omit rapid fluctuations of the input stimulus.}
    \label{fig:refractory_period}
    \vspace*{-0.3cm}
\end{figure}
\subsection{Fast changes in refractory period}
\label{subsec:rate_for_refr}

The refractory period in an event camera influences the duration before a pixel can generate another event, inversely affecting the event rate. Our control strategy uses this relationship to modulate the event rate in real time, targeting an optimal event rate range $[R^-, R^+]$ for enhanced place recognition performance. This is analogous to the refractory period in a biological nerve cell \cite{gracca2023shining, neuronal-dynamics}. As illustrated in Figure \ref{fig:refractory_period}, higher refractory period settings lead to lower event rates in device output.

Let $\mathcal{E}(t, \tau)$ be the set of events that occurred within the time window $[t - \tau, t)$, where $t'$ is the timestamp of the event:
\begin{equation}
    \mathcal{E}(t, \tau) = {\mathbf{e}(x, y, t', p) \in \mathcal{E} \mid t - \tau \leq t' < t}.
    \label{eqn:event_rate}
\end{equation}
Then, the event rate $R(t, \tau)$ (measured in events per second) at time $t$ over the duration $\tau$ is defined as:
\begin{equation}
    R(t, \tau) = \frac{|\mathcal{E}(t, \tau)|}{\tau}.
    \label{eqn:events_per_sec}
\end{equation}

We define a linear mapping $B_\text{r}=f(R)$ from the event rate $R \in [R^-, R^+]$ to the refractory period $B_\text{r} \in [B_\text{r}^-, B_\text{r}^+]$. For the current event rate $R$, we obtain the refractory period bias as:
\begin{equation}
B_\text{r} =
B_\text{r}^+ - \left(\frac{B_\text{r}^+ - B_\text{r}^-}{R^+ - R^-}\right) \cdot (R - R^-).
\label{eqn:fast_changes}
\end{equation}
  Further, we ensure the bias settings remain within the desired bounds [$B_\text{r}^-, B_\text{r}^+]$, resulting in the final bias setting $\hat{B}_\text{r}$: 
\begin{equation}
    \hat{B}_\text{r} = 
    \begin{cases}
    B_\text{r}^+ & \text{if } B_\text{r} > B_\text{r}^+\\
    B_\text{r}^- & \text{if } B_\text{r} < B_\text{r}^-\\
    B_\text{r} & \text{otherwise}.
    \end{cases}
\end{equation}

\subsection{Slow changes in pixel bandwidth and event threshold}
\label{subsec:refr_for_biases}
\begin{figure}[t]
    \centering
    \includegraphics[width=0.49\linewidth]{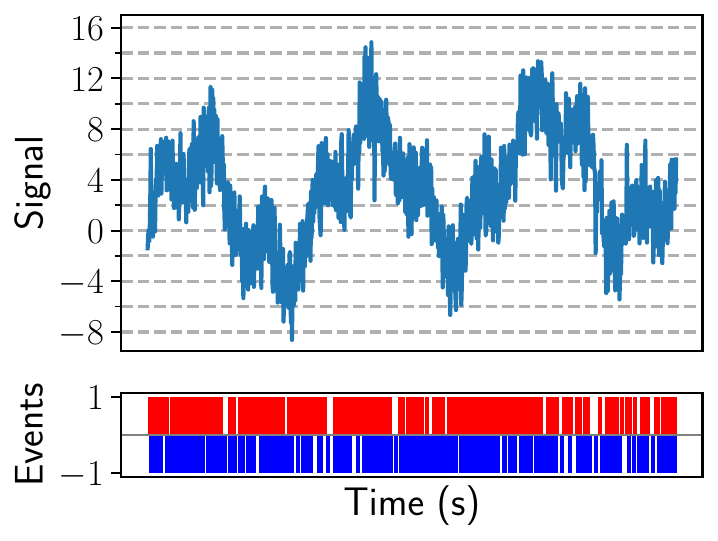}
    \includegraphics[width=0.49\linewidth]{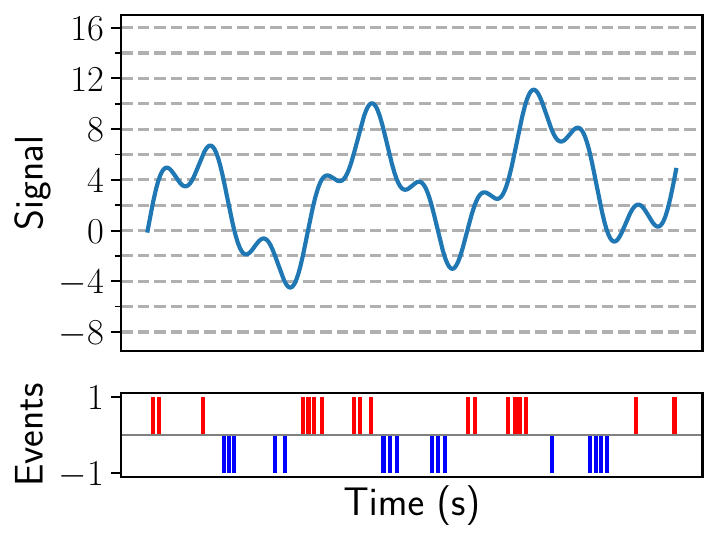}
    \caption{\textbf{Pixel bandwidth.} Left: A high bandwidth allows high frequency signal to pass through, which leads to higher event rates. Right: A low bandwidth filters high frequency components of the signal, leading to a low frequency signal and consequently lower event rate.}
    \label{fig:pixel_bandwidth}
\end{figure}

\begin{figure}[t]
    \centering
    \includegraphics[width=0.49\linewidth]{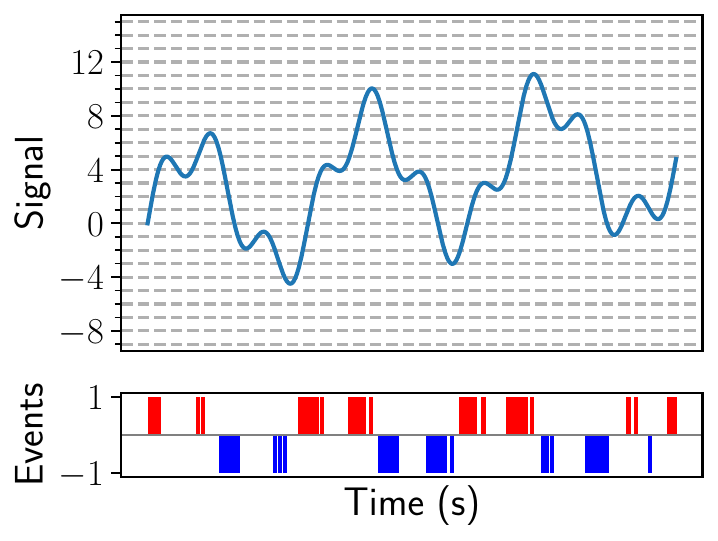}
    \includegraphics[width=0.49\linewidth]{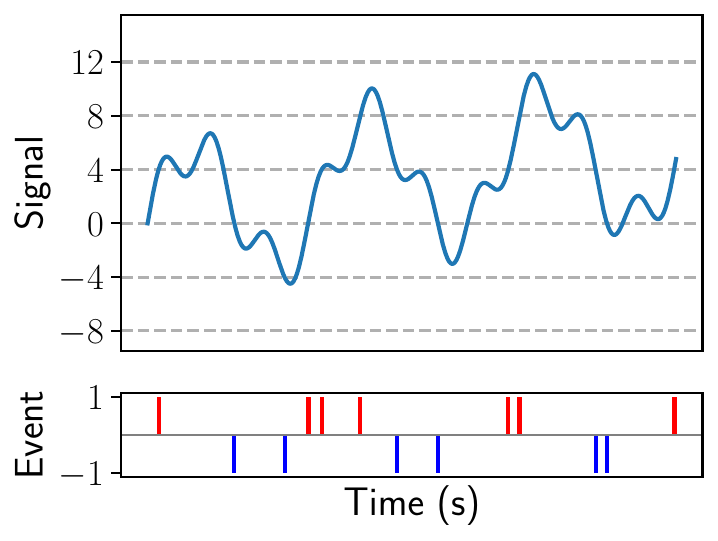}
    \caption{\textbf{Event threshold.} Left: A low event threshold setting (indicated by dashed horizontal lines) results in a high event rate (frequent events in the bottom plot), capturing most changes in the input signal. Right: A high threshold setting results in a lower event rate (much fewer events, as indicated by the sparse red and blue lines), only responding to larger variations in the input signal.}
    \label{fig:event_threshold}
\end{figure}

The pixel bandwidth determines the maximum permitted frequency of input signal, filtering away all higher frequency components \cite{gracca2023shining}. Figure \ref{fig:pixel_bandwidth} illustrates how higher bandwidth increases the output event rate. Event thresholds concern with the minimum change in light intensity required to generate event spikes. Figure \ref{fig:event_threshold} illustrates how higher thresholds lower the output event rate.

We vary the pixel bandwidth and event thresholds to maintain event rates within desirable range $R \in [R^-, R^+]$. However, the instances of these changes have to be monitored because changing pixel bandwidth or event thresholds lead to momentarily disruptive event bursts in the output. Hence slow and simultaneous fixed-step adaptations $\Delta B_\text{p}$ and $\Delta B_\text{e}$ are performed in pixel bandwidth $B_\text{p}$ and event thresholds $B_\text{e}$ respectively if the event rate remains outside the desired bounds for the previous $N$ consecutive time samples despite fast changes in the refractory period (Section~\ref{subsec:rate_for_refr}). Specifically, at time $t$: 
\begin{equation}
    B_\text{p} = 
    \begin{cases}
    B_\text{p} + \Delta B_\text{p} \textit{, if } R_{i} < R_{i}^-\ \forall i \in (t-N, ..., t] \\
    B_\text{p} - \Delta B_\text{p} \textit{, if } R_{i} > R_{i}^+\ \forall i \in (t-N, ..., t],
    \end{cases}
\end{equation}
and similarly for $B_\text{e}$ noting a change in signs:
\begin{equation}
    B_\text{e} = 
    \begin{cases}
    B_\text{e} - \Delta B_\text{e} \textit{, if } R_{i} < R_{i}^-\ \forall i \in (t-N, ..., t] \\
    B_\text{e} + \Delta B_\text{e} \textit{, if } R_{i} > R_{i}^+\ \forall i \in (t-N, ..., t].
    \end{cases}
\end{equation}
\section{Experimental setup}
\label{subsec:setup}
We first describe the robotic platform and event camera processing pipeline in Section~\ref{subsec:robotplatform}. Implementation details are provided in Section~\ref{subsec:implementation}, and the underlying place recognition technique is briefly summarised in Section~\ref{subsec:validation}. Finally, the baseline techniques are described in \mbox{Section~\ref{subsec:baselines}}.
\begin{figure}[t]
    \centering
    \includegraphics[width=0.85\linewidth]{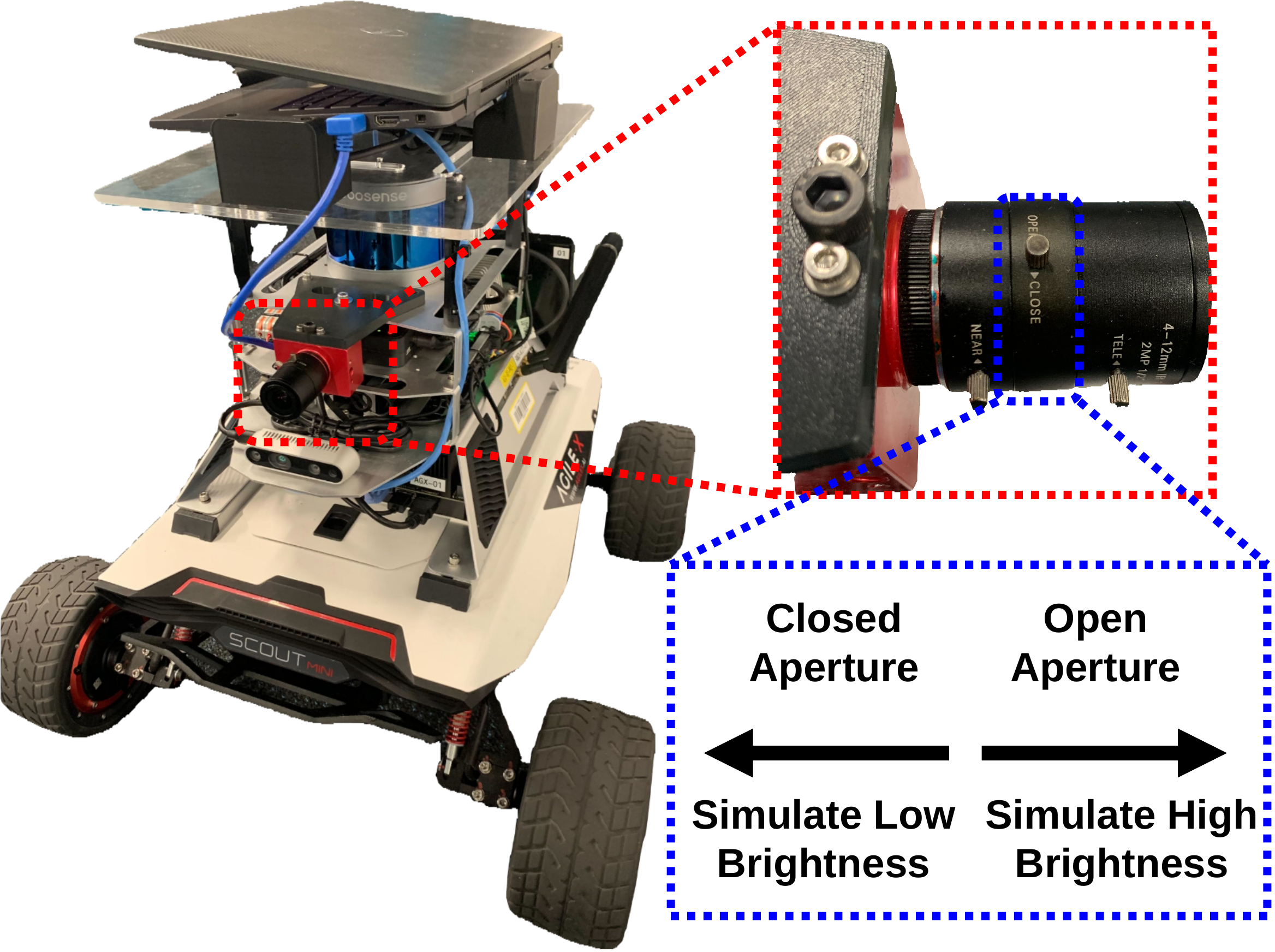}
    \includegraphics[height=0.4\linewidth]{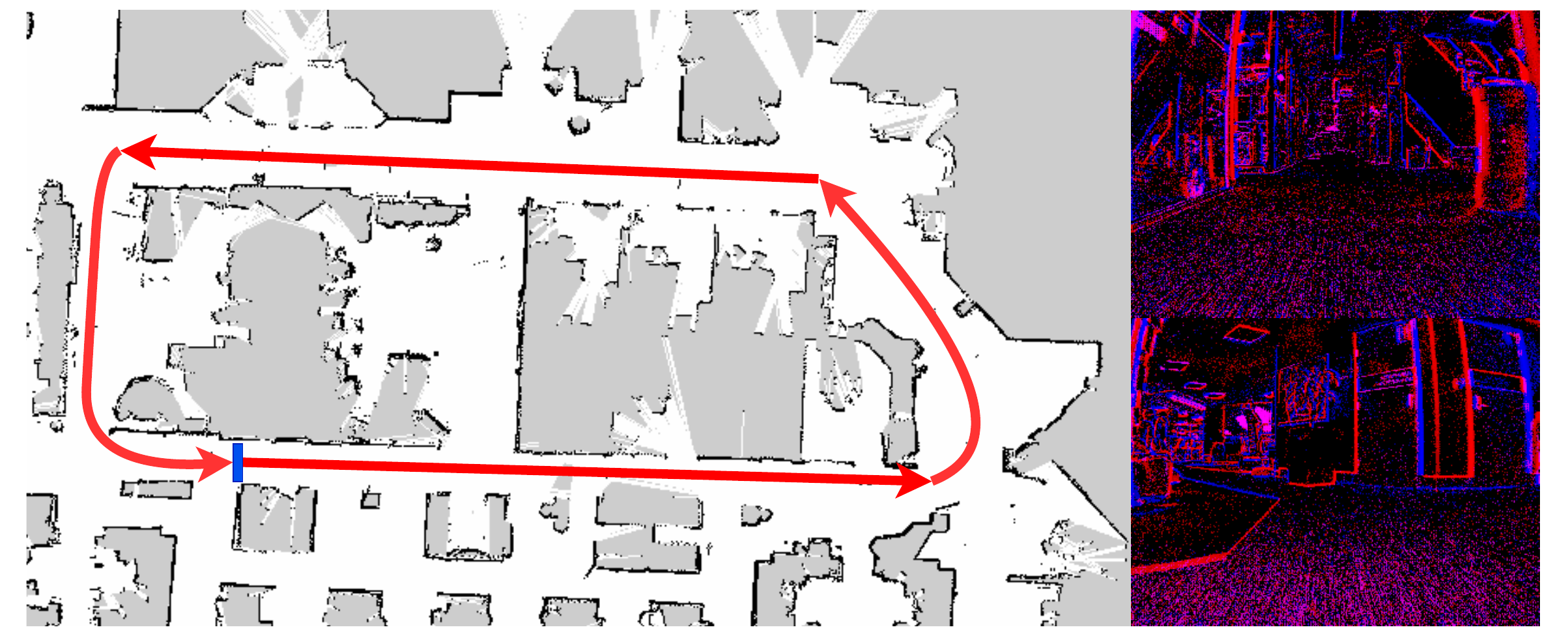}
    \caption{\textbf{(Top) Robot system setup.} Left: Scout Mini platform equipped with various sensors, including a LiDAR to obtain pseudo ground truth and the DAVIS346 event camera mounted forward-facing. Right: Close-up of the DAVIS346 event camera, where the aperture setting can be manually adjusted. Altering the aperture simulates different brightness conditions: closing it simulates low-light environments, while opening it simulates high-brightness scenarios. \textbf{(Bottom) Map from the robot and frames from the event-data}. Left: The blue colored segment marks the starting position and red arrows indicate the traversal path. Right: Frames of accumulated event-data taken from one of the traversals.}
    \vspace*{-0.2cm}
    \label{fig:setup}
\end{figure}
\subsection{Robot platform}
\label{subsec:robotplatform}

We mount an iniVation DAVIS346 event camera \cite{berner2013240} in the front of an AgileX Scout Mini mobile robot such that the camera faces in the direction of motion (Figure~\ref{fig:setup}, left). We ensure that the camera focus remains constant in all our experiments, and use manual changes in camera aperture to simulate various environment brightness conditions (Figure~\ref{fig:setup}, right). While no deliberate changes in the light source positions were introduced, they might have occurred naturally as the dataset was collected over a prolonged time. Although Active Pixel Sensor (APS) grayscale images and Inertial Measurement Unit (IMU) data were also recorded along with the event data, they were not used in the experiments.

Pseudo ground truth robot positions in a fixed world frame are obtained using the SLAM Toolbox ROS package \cite{macenski2021slam}, which runs a full SLAM pipeline that takes the raw odometry from the wheel-encoders and LiDAR data as input to provide accurate ground truth position information in a global coordinate frame in a pre-mapped environment. 

\subsection{Closed-loop bias controller}
\label{subsec:implementation}

Events are streamed over UDP for real-time computation of event rates. Our feedback bias controller uses event rates as feedback to control pixel bandwidth, event thresholds and refractory period in an online manner. Following up on Equation \ref{eqn:event_rate}, the event rates are calculated and updated over time intervals of $\tau = 300ms$.

\begin{table*}
\aboverulesep=0ex
\belowrulesep=0ex
\renewcommand{\arraystretch}{1.2}
\centering
\setlength\tabcolsep{5pt} %
\caption{Recall @ 1 comparison for a reference set recorded in high brightness conditions, with varying brightness conditions for the query set. Baselines include default parameters and three baseline controllers proposed in~\cite{delbruck2021feedback} (PxBw: Control pixel bandwidth, RfPr: Control of refractory period, PxTh: Control pixel threshold; see Section~\ref{subsec:baselines}). Our method consistently outperforms or performs as well as all of the baseline techniques.}
\label{table:primary}
\begin{tabular}{c|c|ccccc|ccccc|ccccc}
\toprule
\multirow{10}{*}{\rotatebox[origin=c]{90}{High Brightness Reference Set}} & & \multicolumn{5}{c|}{Low Brightness (Maximum Variation)} & \multicolumn{5}{c|}{Medium Brightness (Moderate Variation)} & \multicolumn{5}{c}{High Brightness (Minimum Variation)} \\
& & Default & PxBw & RfPr & PxTh & & Default & PxBw & RfPr & PxTh & & Default & PxBw &  RfPr & PxTh & \\
& Run \# & param. & \cite{delbruck2021feedback} & \cite{delbruck2021feedback} & \cite{delbruck2021feedback} & Ours & param. & \cite{delbruck2021feedback} & \cite{delbruck2021feedback} & \cite{delbruck2021feedback} & Ours & param. & \cite{delbruck2021feedback} & \cite{delbruck2021feedback} & \cite{delbruck2021feedback} & Ours \\
\cmidrule(l{3pt}r{3pt}){2-2}\cmidrule(l{3pt}r{3pt}){3-7}\cmidrule(l{3pt}r{3pt}){8-12}\cmidrule(l{3pt}r{3pt}){13-17}
& 1 & 0.28 & 0.39 & 0.40 & 0.75 & 0.77 & 0.98 & 0.97 & 0.94 & 0.91 & 0.96 & 0.99 & 0.97 & 0.93 & 0.92 & 0.99 \\
& 2 & 0.47 & 0.71 & 0.62 & 0.86 & 0.79 & 0.88 & 0.98 & 0.98 & 0.93 & 0.96 & 0.95 & 0.95 & 0.97 & 0.95 & 0.94 \\
& 3 & 0.52 & 0.67 & 0.45 & 0.76 & 0.92 & 0.96 & 0.98 & 0.87 & 0.90 & 0.99 & 0.94 & 0.97 & 0.89 & 0.93 & 0.97 \\
& 4 & 0.44 & 0.33 & 0.49 & 0.84 & 0.90 & 0.98 & 0.95 & 0.85 & 0.91 & 0.92 & 0.97 & 0.96 & 0.85 & 0.86 & 0.93 \\
& 5 & 0.46 & 0.44 & 0.46 & 0.91 & 0.89 & 0.98 & 0.91 & 0.89 & 0.92 & 0.97 & 0.96 & 0.95 & 0.70 & 0.96 & 0.97 \\
\cmidrule(l{3pt}r{3pt}){2-2}\cmidrule(l{3pt}r{3pt}){3-7}\cmidrule(l{3pt}r{3pt}){8-12}\cmidrule(l{3pt}r{3pt}){13-17}
& Average & 0.43 & 0.51 & 0.48 & 0.83 & \textbf{0.85} & \textbf{0.96} & \textbf{0.96} & 0.91 & 0.91 & \textbf{0.96} & \textbf{0.96} & \textbf{0.96} & 0.87 & 0.92 & \textbf{0.96} \\
& Std.~Dev. & 0.09 & 0.17 & 0.08 & 0.07 & 0.07 & 0.04 & 0.03 & 0.05 & 0.01 & 0.03 & 0.02 & 0.01 & 0.10 & 0.04 & 0.02 \\
\bottomrule
\end{tabular}
\end{table*}
\begin{figure*}[ht]
    \centering
    \includegraphics[width=0.6\linewidth]{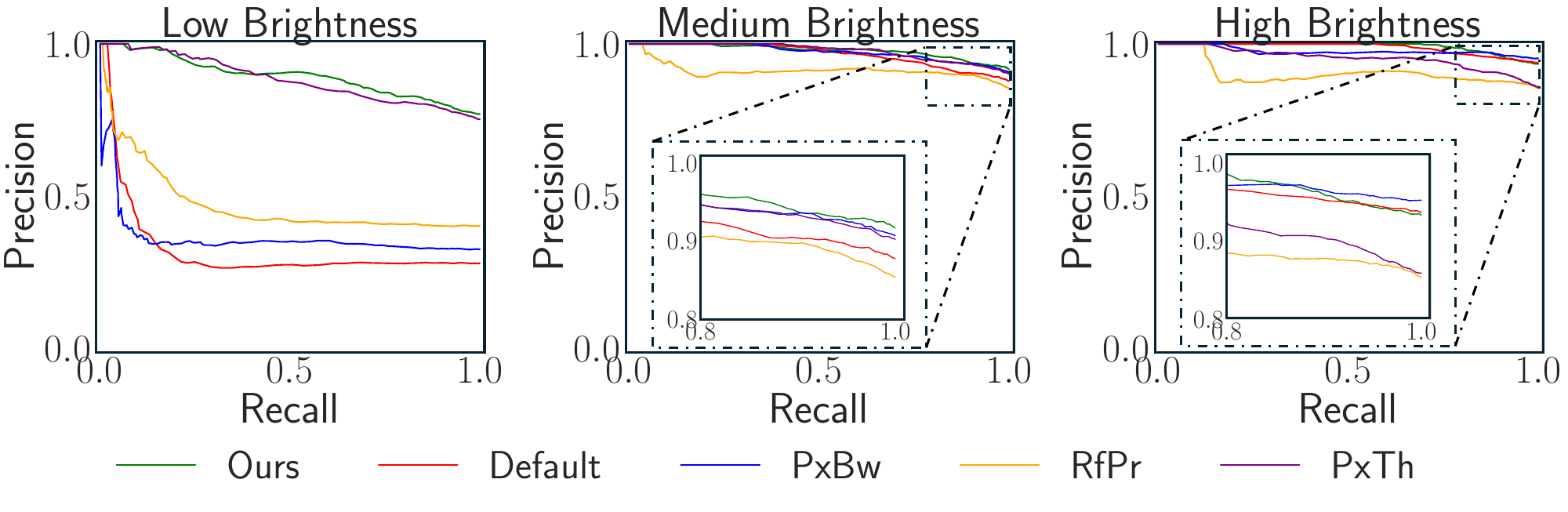}
    \vspace*{-0.3cm}
    \caption{\textbf{Precision-Recall Curves for worst case performance (reference: high brightness).} The worst performing run from each method is chosen for plotting the PR curves. While our method showcases superior performance in comparison to all baselines when the query set is taken from low and medium brightness conditions, its performance remains on-par with the top three approaches when query set is from high brightness conditions.}
    \label{fig:pr_curve_high_ref}
    \vspace*{-0.1cm}
\end{figure*}
\subsubsection{Bias current initializations (in Amperes)}
\label{implementation_init}

Initial pixel bandwidth is set through photoreceptor bias $B_\text{pr} = 44.16pA$ and source-follower bias $B_\text{sf} = 1.48pA$. Initial event thresholds are set as ON threshold $B_\text{on} = 762.89nA$ and OFF threshold $B_\text{off} = 498.62pA$.

\subsubsection{Fast changes in refractory period}
\label{implementation_fast_changes}

As explained in Section \ref{subsec:rate_for_refr}, we maintain a desired range of event rate $R \in [R^-, R^+]$ where $R^- = 5 \times 10^5 Hz$ and $R^+ = 2.5 \times 10^6 Hz$. We limit variations in refractory period to $B_\text{r} \in [759.37pA,10.25nA]$.

\subsubsection{Slow changes in pixel bandwidth and event threshold}
\label{implementation_slow_changes}

Following the explanation in section \ref{subsec:refr_for_biases}, we maintain the same desired range for event rate $R \in [R^-, R^+]$ as in Section~\ref{implementation_fast_changes}. Slow changes are performed if $R \notin [R^-, R^+]$ for the last $N = 5$ consecutive time steps. Variations in pixel bandwidth $\Delta B_\text{p} = (\Delta B_\text{pr}, \Delta B_\text{sf})$ is performed as fixed step updates in photoreceptor bias $\Delta B_\text{pr}=1.85pA$ and source-follower bias $\Delta B_\text{sf}=0.23pA$. Similarly, variations in event threshold $\Delta B_\text{e} = (\Delta B_\text{on}, \Delta B_\text{off})$ is performed through fixed step updates in ON threshold $\Delta B_\text{on}=60.55nA$ and OFF threshold $\Delta B_\text{off}=14.84pA$.

\subsection{Visual Place Recognition Pipeline}
\label{subsec:validation}
Event frames~\cite{rebecq2016evo, maqueda2018event} are generated by accumulating events into non-overlapping fixed-sized time-windows of $66ms$ duration. In brief, each pixel of the event frame contains the number of events that occurred within the time window. Post-processing involves filtering out event frames with event bursts and filtering out hot pixels, which produce noise in the form of high-frequency event spikes which are not caused by changes in light intensity \cite{hu2021v2e}. Additionally, frames where the robot is stationary are discarded.

\begin{table*}
\aboverulesep=0ex
\belowrulesep=0ex
\setlength\tabcolsep{5pt} %
\centering
\renewcommand{\arraystretch}{1.2}
\caption{Recall @1 comparison for a reference set in low brightness conditions, with varying brightness conditions for the query set. See Table~\ref{table:primary} for baseline techniques.}
\label{table:secondary}
\begin{tabular}{c|c|ccccc|ccccc|ccccc}
\toprule
\multirow{10}{*}{\rotatebox[origin=c]{90}{Low Brightness Reference Set}} & & \multicolumn{5}{c|}{Low Brightness (Minimum Variation)} & \multicolumn{5}{c|}{Medium Brightness (Moderate Variation)} & \multicolumn{5}{c}{High Brightness (Maximum Variation)} \\
& & Default & PxBw & RfPr & PxTh & & Default & PxBw & RfPr & PxTh & & Default & PxBw &  RfPr & PxTh & \\
& Run \# & param. & \cite{delbruck2021feedback} & \cite{delbruck2021feedback} & \cite{delbruck2021feedback} & Ours & param. & \cite{delbruck2021feedback} & \cite{delbruck2021feedback} & \cite{delbruck2021feedback} & Ours & param. & \cite{delbruck2021feedback} & \cite{delbruck2021feedback} & \cite{delbruck2021feedback} & Ours \\
\cmidrule(l{3pt}r{3pt}){2-2}\cmidrule(l{3pt}r{3pt}){3-7}\cmidrule(l{3pt}r{3pt}){8-12}\cmidrule(l{3pt}r{3pt}){13-17}
& 1 & 0.88 & 0.92 & 0.81 & 0.84 & 0.96 & 0.47 & 0.73 & 0.60 & 0.82 & 0.92 & 0.59 & 0.73 & 0.62 & 0.74 & 0.89 \\
& 2 & 0.88 & 0.99 & 0.83 & 0.91 & 0.92 & 0.73 & 0.97 & 0.81 & 0.60 & 0.91 & 0.73 & 0.94 & 0.80 & 0.56 & 0.97 \\
& 3 & 0.96 & 0.98 & 0.87 & 0.90 & 0.97 & 0.84 & 0.96 & 0.75 & 0.92 & 0.93 & 0.80 & 0.96 & 0.67 & 0.91 & 0.85 \\
& 4 & 0.85 & 0.94 & 0.72 & 0.94 & 0.97 & 0.61 & 0.92 & 0.47 & 0.80 & 0.91 & 0.65 & 0.92 & 0.59 & 0.78 & 0.88 \\
& 5 & 0.95 & 0.99 & 0.83 & 0.95 & 0.97 & 0.85 & 0.81 & 0.46 & 0.67 & 0.96 & 0.60 & 0.82 & 0.65 & 0.78 & 0.94 \\
\cmidrule(l{3pt}r{3pt}){2-2}\cmidrule(l{3pt}r{3pt}){3-7}\cmidrule(l{3pt}r{3pt}){8-12}\cmidrule(l{3pt}r{3pt}){13-17}
& Average & 0.90 & \textbf{0.96} & 0.81 & 0.91 & \textbf{0.96} & 0.70 & 0.88 & 0.62 & 0.76 & \textbf{0.93} & 0.67 & 0.87 & 0.67 & 0.75 & \textbf{0.91} \\
& Std.~Dev. & 0.05 & 0.03 & 0.05 & 0.04 & 0.02 & 0.16 & 0.11 & 0.16 & 0.13 & 0.02 & 0.09 & 0.10 & 0.08 & 0.13 & 0.05 \\
\bottomrule
\end{tabular}
\end{table*}
\begin{figure*}[t]
    \centering
    \includegraphics[width=0.6\linewidth]{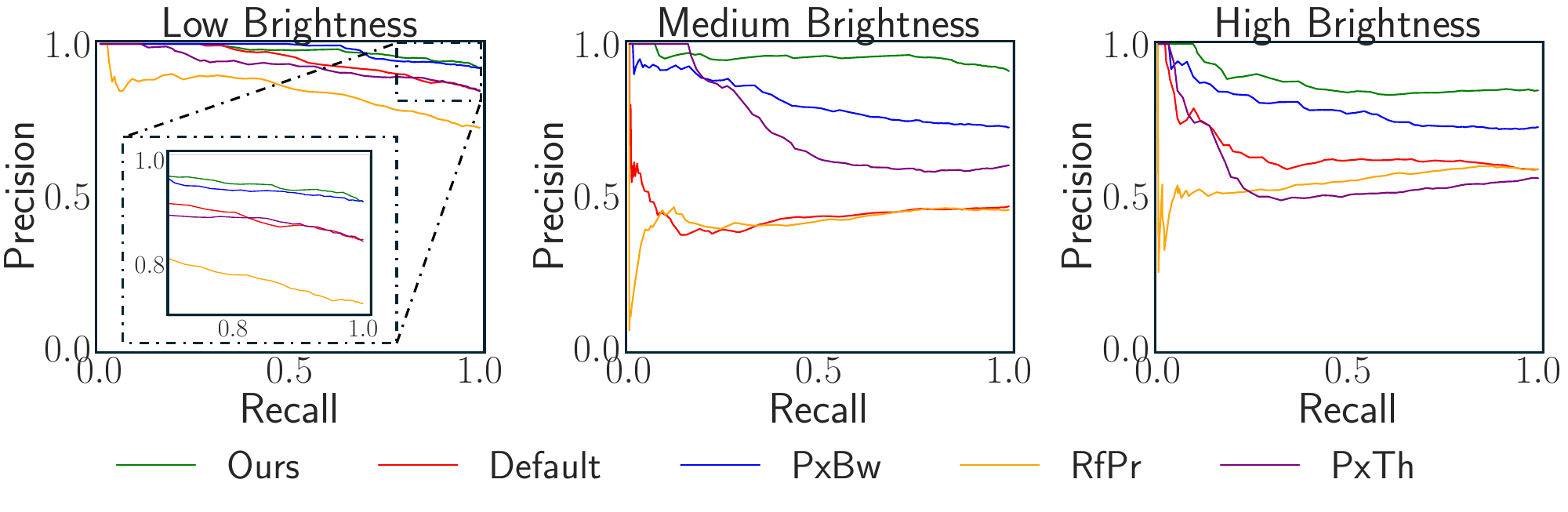}
    \vspace*{-0.3cm}
    \caption{\textbf{Precision-Recall Curves for worst case performance (reference: low brightness).} The worst performing run from each method is chosen for plotting the PR curves. The PR curves demonstrate that our fast-and-slow bias controller performs better than all baseline techniques when considering the worst-case performance.}
    \label{fig:pr_curve_low_ref}
    \vspace*{-0.3cm}
\end{figure*}
We use a previously proposed, simple visual place recognition pipeline to demonstrate our feedback controller \cite{milford2015towards}. To ensure robustness against appearance changes, the event frames are patch-normalized~\cite{milford2012seqslam} with a patch-size of $8px$. Sum-of-absolute-differences (SAD) is performed between event frames to obtain distance matrices -- the higher the SAD, the less similar the two event frames~\cite{milford2012seqslam}. The event frame in the reference dataset that has the lowest SAD to the current query event frame is considered as the top match.

These top matches are compared with the ground truth robot positions in the map with a tolerance of $5m$. Recall $R$ and precision $P$~\cite{schubert2023visual} are computed for each experiment as: $R=\frac{TP}{Q}$ and $P=\frac{TP}{TP+FP}$, where $TP$ indicates the number of true positives, $FP$ indicates the number of false positives and $Q$ indicates the number of query images. The Recall@1 (R@1) metric is identical to the precision at 100\% recall.

\subsection{Baselines}
\label{subsec:baselines}

\subsubsection{Default parameters}
\label{subsubsec:default_baseline}

The first baseline technique is to naively use the default biases that are used by jAER when it is started \cite{delbruck2008frame}. This corresponds to the use case where no manual and/or automated tuning is performed at all. %

\subsubsection{Feedback control of refractory period to limit event rate}
\label{subsubsec:tobi_refr_baseline}
The second baseline technique \cite{delbruck2021feedback} is similar to our fast feedback controller and adapts the refractory period only. As a recap, a higher refractory periods reduces the event rate and vice versa. Specifically, the controller aims to keep event rate below a predefined upper bound by reducing the refractory period before returning to the system default. Note that this controller does not lead to spurious event bursts as only the refractory period is changed. Our \textit{fast} adaptation method diverges from the above approach by tightly constraining the adaptation of the refractory period to a narrower range. 

\subsubsection{Feedback control of pixel bandwidth to limit noise}
\label{subsubsec:tobi_bandwidth_baseline}

The output event rate tends to increase with an increase in pixel bandwidth. The noise event rate also increases with an increase in pixel bandwidth. This third baseline \cite{delbruck2021feedback} limits the event noise rate in the output data by regulating the pixel bandwidth. Furthermore, the noise events are separated from the signal events using a background activity (BA) filter~\cite{hu2021v2e}. In contrast, our \textit{slow} adaptation method performs simultaneous adjustments to pixel bandwidth and event threshold parameters on a case-by-case basis while monitoring the event rate.

\subsubsection{Feedback control of event threshold to bound noise}
\label{subsubsec:tobi_threshold_baseline}

The event threshold determines how much of a change is required in log input light intensity for an event spike to occur. The fourth baseline \cite{delbruck2021feedback} exploits event thresholds to maintain event rates within desirable bounds. Unlike refractory period control, changes in the event threshold and pixel bandwidth lead to event bursts, making the device's output momentarily noisy. In contrast, similar to Section \ref{subsubsec:tobi_bandwidth_baseline}, our \textit{slow} adaptation approach performs sparse and simultaneous changes in pixel bandwidth and event threshold parameters while monitoring the event rate.
\section{Results}
\label{sec:results}

Section \ref{subsec:quantitative} compares our closed-loop bias controller's performance against the baseline techniques, with the main results provided in Tables \ref{table:primary} and \ref{table:secondary}. We also perform ablations studies on the optimal frequency of performing slow changes in Section \ref{subsubsec:slow_changes}, and compare and analyse individual contributions of various decoupled components of our closed-loop bias controller pipeline in Section \ref{subsubsec:components}.

\subsection{Comparison to state-of-the-art}
\label{subsec:quantitative}
Tables~\ref{table:primary} and~\ref{table:secondary} provide the Recall@1 metrics for varying query brightness conditions (low, medium, and high brightness) when the reference set was recorded under high brightness (Table~\ref{table:primary}) or low brightness (Table~\ref{table:secondary}). Each experiment was performed with five repetitions. Generally speaking, in all conditions our method outperforms all baseline techniques or performs at least as well as the best performing baseline, and it performs the most consistently.

Specifically, in Table~\ref{table:primary} (reference dataset: high brightness) one can observe that while the default bias parameters and the pixel bandwidth controller proposed in~\cite{delbruck2021feedback} perform as well as our technique in medium and high brightness conditions (all R@1=0.96), they perform considerably worse in the challenging low query brightness conditions (absolute R@1 improvement of 42\% and 34\% respectively). This indicates that our proposed controller is more robust when the appearance changes significantly (i.e.~reference in high brightness and query in low brightness). We note that the pixel threshold controller consistently performs 2-5\% worse than our proposed fast-and-slow controller. Another notable observation is that the lowest performance across all repetitions for our technique is almost always higher than the lowest performance of all baseline techniques, with the single exception being repetition 4 in the high brightness condition. This is further shown in Figure~\ref{fig:pr_curve_high_ref}.

The same trend can be observed in Table~\ref{table:secondary} (reference dataset: low brightness), where the performance improvement over the second-best-performing pixel bandwidth controller~\cite{delbruck2021feedback} is an absolute 4\% in the high brightness query condition, the query condition with the highest appearance change. Figure~\ref{fig:pr_curve_low_ref} demonstrates that the performance advantage in the worst-case run becomes even more pronounced, comparing a R@1 of 85\% of our technique with 59\% (default parameters), 73\% (pixel bandwidth controller), 59\% (refractory period controller) and 56\% (pixel threshold controller).

\subsection{Ablation Studies}
\label{subsec:ablation}

\subsubsection{Frequency of slow changes}
\label{subsubsec:slow_changes}
The pixel bandwidth and event threshold bias settings are adjusted if the observed event rate remains outside desired range for the previous $N$ times (Section~\ref{subsec:refr_for_biases}). Larger $N$ lead to fewer number of slow changes made in a traversal, which is preferable since changes in pixel bandwidth and event threshold lead to event bursts. Filtering out events due to event bursts leads to a loss of useful data; and in the worst case, loss of localization. However, not tuning the pixel bandwidth and event threshold would also be destructive as those changes are required to keep the event rate within desired bounds (Figure~\ref{fig:APSvariations1}).

We analyze the place recognition performance when running experiments with $N = \{2, 5, 7, 10\}$. Figure \ref{fig:ablations} (left) illustrates the average R@1 values obtained for three runs of each experiment. The most consistent performance is obtained when $N^*=5$, including the highest performance in low-brightness query sets.
\subsubsection{Components}
\label{subsubsec:components}

We analyze individual components in our bias control pipeline to understand their contribution toward the overall pipeline performance. The variations are 1) to only initialize the event camera with the parameters outlined in Section~\ref{implementation_init} without changing the bias parameters during the run (constant), 2) only performing `fast' changes in the refractory period, 3) only performing `slow' adaptations in pixel bandwidth and event threshold, and 4) performing both slow and fast changes in refractory period.

\begin{figure}[t]
    \centering
    \includegraphics[width=0.49\linewidth]{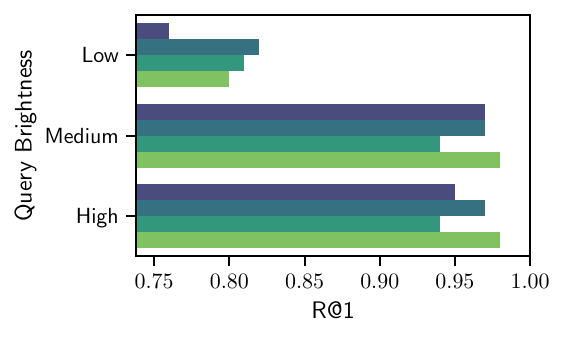}
    \includegraphics[width=0.49\linewidth]{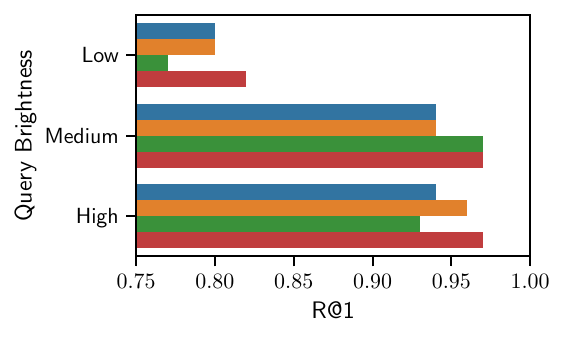}\\
    \hspace{0.12\linewidth}
    \includegraphics[height=0.05\linewidth]{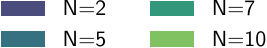}
    \hspace{0.15\linewidth}
    \includegraphics[height=0.05\linewidth]{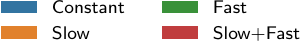}
    \caption{\textbf{Ablation studies. Left: }We investigate the frequency of slow changes. Generally, the performance differences are rather small, indicating good robustness of our technique to this hyperparameter. \textbf{Right: }Contributions of individual components of our online bias controller. The combination of slow+fast leads to the highest performance in all cases.}
    \label{fig:ablations}
    \vspace*{-0.3cm}
\end{figure}

Figure \ref{fig:ablations} (right) shows the R@1 performance from three iterations of each variation. When the query set is recorded in low brightness conditions, all variants significantly improve upon the constant baseline technique. In the medium and high brightness query conditions, the performance differences are relatively small; with our fast+slow controller achieving the highest performance in these conditions. Interestingly, only changing the refractory period (`fast') leads to a small performance decrease compared to the constant baseline. 

\section{Discussion and Conclusions}
\label{sec:discussions}

This paper has introduced a novel feedback control approach for event cameras, integrating fast-and-slow mechanisms to dynamically adjust bias parameters, thereby optimizing event stream processing for VPR tasks. Our methodology not only enhances the adaptability of event cameras to varying environmental conditions but also demonstrates that a feedback controller built specifically for the downstream task of VPR outperforms a general feedback controller in the same downstream task.
Future work will extend this approach to longer trajectories and larger environmental variations, such as day-night transitions and diverse weather conditions. We also plan to integrate our feedback control system with Spiking Neural Networks (SNNs) \cite{hussaini2023ensembles, hines2023vprtempo, lee2023ev, yu2023brain, zhu2023neuromorphic}. SNNs, with their bio-inspired processing capabilities, could significantly benefit from the refined event streams produced by our method, potentially leading to more robust and efficient visual recognition systems.

Incorporating velocity signals from inertial measurement units (IMUs) to dynamically adjust the temporal windows in the VPR pipeline presents another pathway that could deal with the variability of motion in event-based vision. Unlike fixed temporal intervals, velocity-informed event frames could adjust the processing of event streams in real-time, better aligning with the movement dynamics of the scene or the observer. This concept could also extend to bias parameter adjustments, where velocity data could inform the tuning of event camera sensitivities, potentially improving the camera's responsiveness to motion-induced event generation.

Another intriguing direction for future research is the incorporation of environmental light intensity measurements to refine the tuning of event camera biases. This method promises a finer, more context-sensitive calibration of camera settings, achieving an ideal equilibrium in event generation across various lighting scenarios. However, this strategy may necessitate a more sophisticated approach, as relying solely on raw light intensity might be overly simplistic. Incorporating some level of scene comprehension could offer a more nuanced optimization, potentially leading to an improvement in the signal-to-noise ratio and thereby enhancing the quality of the the event stream for downstream tasks.

In conclusion, our fast-and-slow feedback controller represents a significant step forward in the practical application of event cameras for VPR and potentially other robotic vision tasks. By addressing the dynamic nature of environmental conditions and the inherent limitations of current event camera technology, we have laid the groundwork for more adaptive, efficient, and effective event-based vision systems.

\bibliography{bibfiles/references}
\bibliographystyle{styles/IEEEtran}

\end{document}